\documentclass[11pt]{article}
\usepackage{acl2016}
\usepackage{times}
\usepackage{latexsym}

\usepackage[pdftex]{graphicx}
\usepackage{amsmath}
\usepackage{url}

\newtheorem{theorem}{Theorem}
\def\den#1{[\![#1]\!]}
\DeclareMathOperator{\tr}{tr}

\aclfinalcopy 

\title{Learning Semantically and Additively Compositional\\ Distributional Representations}

 \author{Ran Tian \and Naoaki Okazaki \and Kentaro Inui \\
         Tohoku University, Japan \\ {\tt \{tianran, okazaki, inui\}@ecei.tohoku.ac.jp}}

\date{}

\begin{document}

\maketitle

\begin{abstract}
This paper connects a vector-based composition model to a formal semantics, the Dependency-based Compositional Semantics (DCS). We show theoretical evidence 
that the vector compositions in our model conform to the logic of DCS. Experimentally, 
we show that vector-based composition brings a strong ability to calculate similar phrases as similar 
vectors, achieving near state-of-the-art on a wide range of phrase similarity tasks and relation classification; 
meanwhile, DCS can guide building vectors for structured queries that can be directly executed. 
We evaluate this utility on sentence 
completion task and report a new state-of-the-art. 
\end{abstract}

\section{Introduction}
\label{sec:intro}

\begin{figure}[t]
\includegraphics[scale=0.26,bb=20 10 860 660,clip]{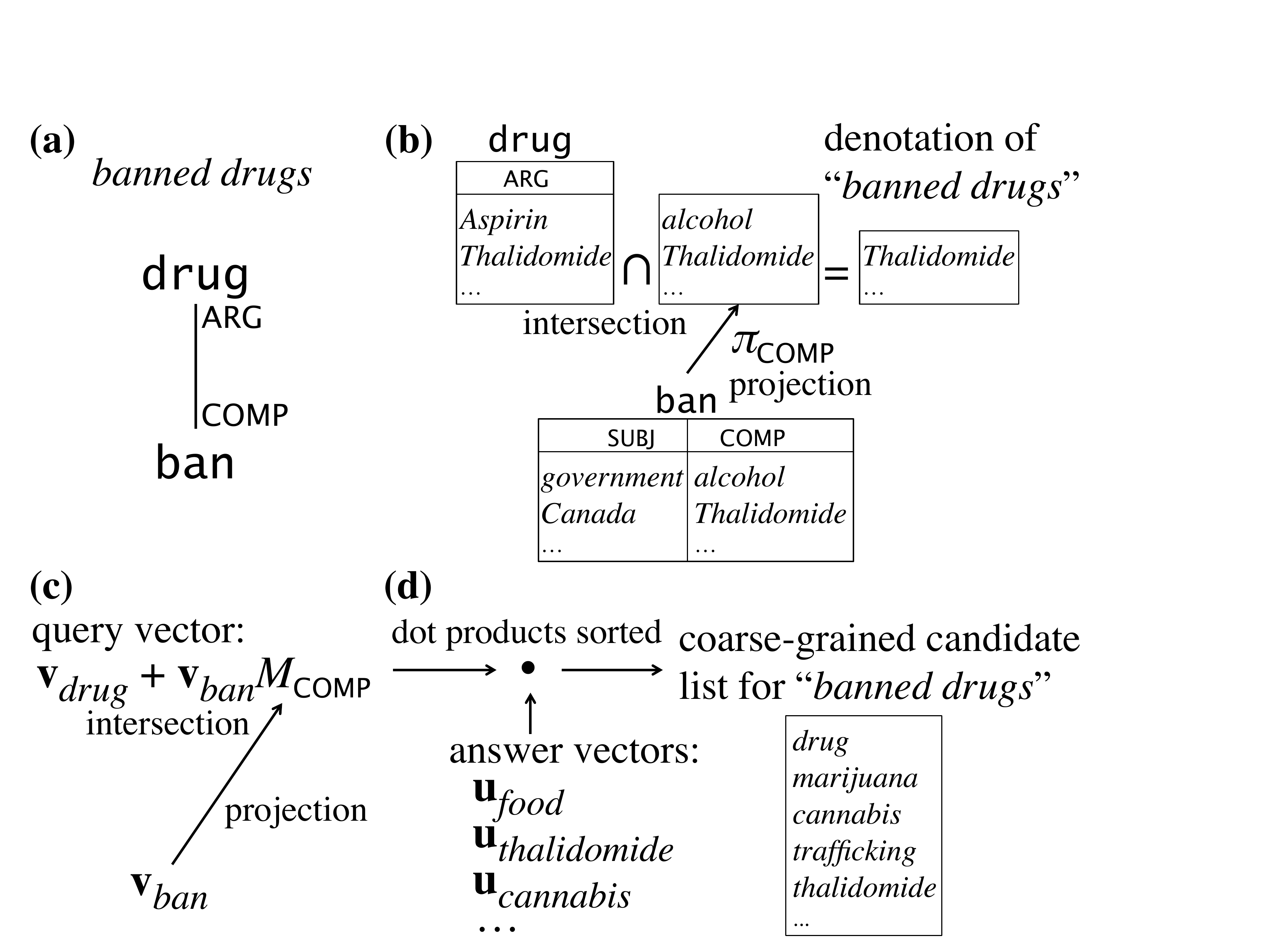}
\caption{(a) The DCS tree of ``\textit{banned drugs}'', which controls (b) the calculation of its denotation. 
In this paper, we learn word vectors and matrices such that (c) the same calculation is 
realized in distributional semantics. 
The constructed query vector can be used to (d) 
retrieve a list of coarse-grained candidate answers to that query.}
\label{fig:dcstree}
\end{figure}

A major goal of semantic processing is to map natural language utterances to representations that facilitate 
calculation of meanings, execution of commands, and/or inference of knowledge. 
Formal semantics supports such representations by defining words as some functional units and 
combining them via a specific logic. A simple and illustrative example is the 
Dependency-based Compositional Semantics (DCS) \cite{liang13}. DCS composes meanings 
%
%
%
from denotations 
of words (i.e.~sets of things to which the words apply); say, the 
denotations of the 
concept \texttt{drug} and the event \texttt{ban} is shown in Figure~\ref{fig:dcstree}b, where \texttt{drug} 
is a list of drug names and 
\texttt{ban} is a list of the subject-complement pairs in any \textit{ban} event; then, a list of \textit{banned drugs} can 
be constructed by first taking the \textsf{COMP} column of all records in 
\texttt{ban} (projection ``$\pi_{\textsf{COMP}}$''), and then intersecting the results with \texttt{drug} 
(intersection ``$\cap$''). This procedure defined how words can be combined to form a meaning. 
Better yet, the procedure can be concisely illustrated by 
the DCS tree of ``\textit{banned drugs}'' (Figure~\ref{fig:dcstree}a), which is similar to a 
dependency tree but 
possesses precise procedural and logical meaning 
(Section~\ref{sec:background}). DCS has been 
shown useful in question answering \cite{liang13} and textual entailment recognition \cite{tian14}.

Orthogonal to the formal semantics of DCS, 
distributional vector representations are useful in capturing lexical semantics of 
words \cite{Turney:2010,levyTACL}, and progress is made in combining the word vectors to form meanings of 
phrases/sentences 
\cite{mitchell10,baroni-zamparelli10,grefenstette-sadrzadeh:2011:EMNLP,socher12,paperno-pham-baroni14,hashimoto-EtAl:2014:EMNLP2014}. 
However, less effort is devoted to finding a link between vector-based compositions and the composition 
operations in any formal semantics. 
We believe that if a link can be found, then symbolic formulas in the formal semantics will 
be realized by vectors composed from word embeddings, such that similar things are realized 
by similar vectors; meanwhile, vectors will acquire formal meanings that can directly be used 
in execution or inference process. 
Still, to find a link is challenging because any vector compositions that realize such 
a link must conform to the logic of the formal semantics. 


In this paper, we establish a link between DCS and certain vector compositions, 
%
%
%
%
%
%
%
achieving a \emph{vector-based DCS} by replacing denotations of words with word vectors, and 
realizing the composition operations such as intersection and projection as addition and linear 
mapping, respectively. For example, to construct a vector for ``\textit{banned drugs}'', one takes 
the word vector $\mathbf{v}_{\textit{ban}}$ and multiply it by a matrix $M_{\textsf{COMP}}$, 
corresponding to the projection $\pi_{\textsf{COMP}}$; then, one adds the result to the word vector 
$\mathbf{v}_{\textit{drug}}$ to realize the intersection operation (Figure~\ref{fig:dcstree}c). 
We provide a method to train the 
word vectors and linear mappings (i.e.~matrices) jointly from unlabeled corpora. 

The rationale for our model is as follows. First, recent research has shown that additive composition 
of word vectors is an approximation to the situation where two words have 
overlapping context \cite{addcomp}; therefore, it is suitable to 
implement an ``and'' or intersection 
operation (Section~\ref{sec:vector}). 
We design our model such that the resulted distributional representations are expected to have 
additive compositionality. Second, 
when intersection is realized as addition, it is natural to implement projection as linear mapping, as 
suggested by the logical interactions between the two operations (Section~\ref{sec:vector}). 
Experimentally, we show that vectors and matrices learned 
by our model exhibit favorable characteristics as compared with vectors trained by GloVe 
\cite{pennington-socher-manning:2014:EMNLP2014} or those learned from syntactic dependencies 
(Section~\ref{sec:quality}). 
Finally, additive composition brings our model a strong ability to 
calculate similar vectors for similar phrases, whereas syntactic-semantic roles 
(e.g.~\textsf{SUBJ}, \textsf{COMP}) can be distinguished by different projection matrices 
(e.g.~$M_{\textsf{SUBJ}}$, $M_{\textsf{COMP}}$). 
We achieve near state-of-the-art performance on a wide range of phrase similarity tasks 
(Section~\ref{sec:phrasesim}) and relation classification (Section~\ref{sec:relclass}). 

Furthermore, we show that a vector as constructed above for 
``\textit{banned drugs}'' can be used as 
a \emph{query vector} to retrieve a coarse-grained candidate list of banned drugs, 
by sorting its dot products with \emph{answer vectors} that are 
also learned by our model (Figure~\ref{fig:dcstree}d). This is due to the ability of our approach to 
provide a language model that can find likely words to fill in the blanks such as 
``\textit{$\rule{2ex}{1pt}$ is a banned drug}'' or 
``\textit{the drug $\rule{2ex}{1pt}$ is banned by \ldots}''. 
A highlight is the calculation being done as if a query is ``executed''  by the 
DCS tree of ``\textit{banned drugs}''. 
We quantitatively evaluate this utility on sentence completion task \cite{zweig-EtAl:2012:ACL2012} 
and report a new state-of-the-art (Section~\ref{sec:sentcomp}).

\section{DCS Trees}
\label{sec:background}

DCS composes meanings from denotations, or sets of things to which words apply. 
A ``thing'' (i.e.~element of a denotation) is represented by a tuple of features of the form \textsf{Field}=\textit{Value}, 
with a fixed inventory of fields. 
For example, a denotation \texttt{ban} might be a set of tuples 
$\texttt{ban}=\{(\textsf{SUBJ}{=}\textit{Canada},\textsf{COMP}{=}\textit{Thalidomide}),\ldots\}$, 
in which each tuple records participants of a banning event (e.g.~Canada banning Thalidomide). 

\begin{figure}[t]
\centering
\includegraphics[scale=0.26,bb=0 0 640 230,clip]{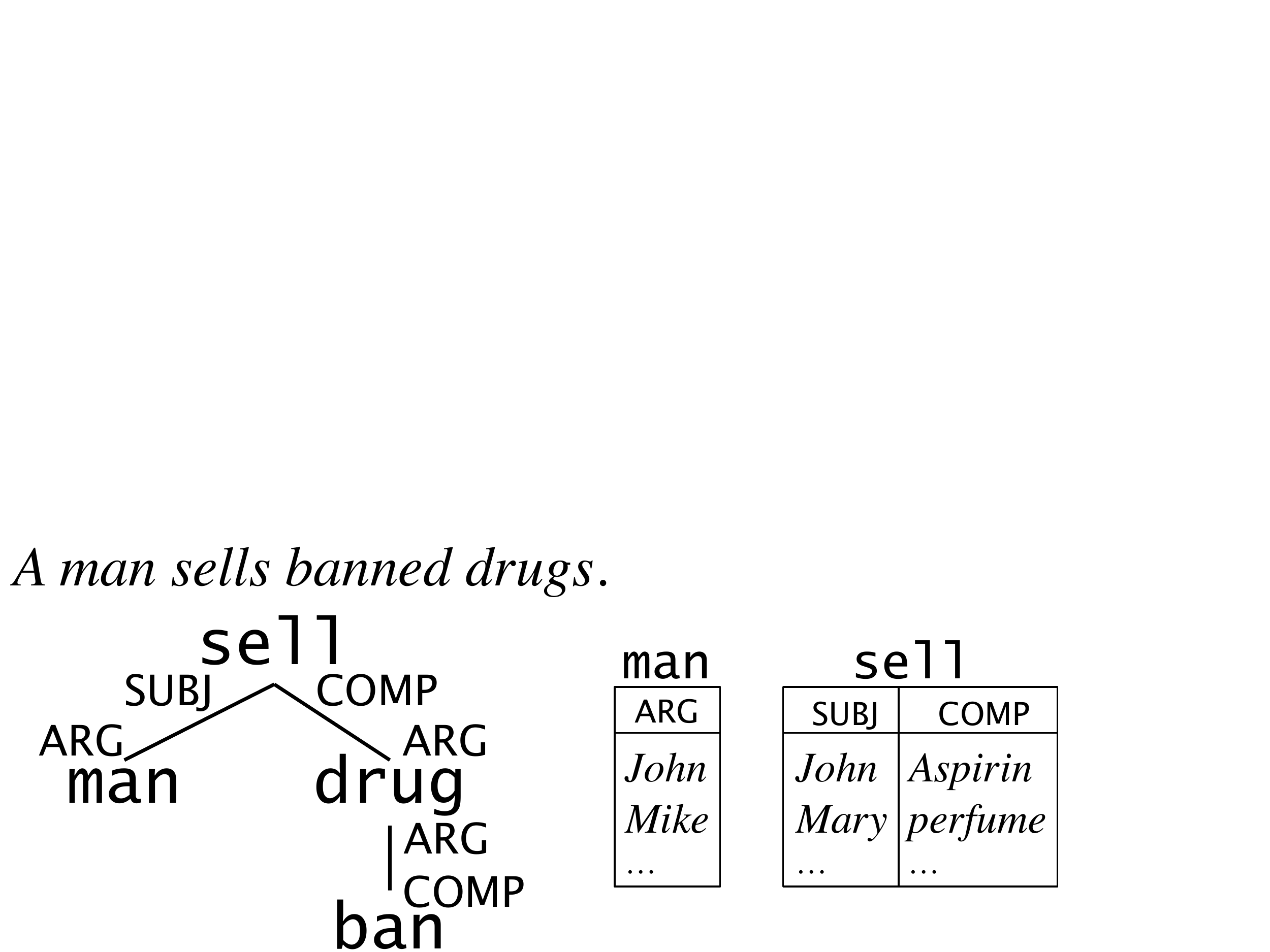}
\caption{DCS tree for a sentence}
\label{fig:dcswatch}
\end{figure}

Operations are applied to sets of things to generate new denotations, for modeling semantic 
composition. An example is the intersection of \texttt{pet} and \texttt{fish} giving the 
denotation of ``\textit{pet fish}''. Another necessary operation is projection; by $\pi_{\textsf{N}}$ we mean 
a function mapping a tuple to its value of the field \textsf{N}. For example, 
$\pi_{\textsf{COMP}}(\texttt{ban})$ is the value set of the \textsf{COMP} fields in \texttt{ban}, which 
consists of banned objects (i.e.~$\{\textit{Thalidomide},\ldots\}$). 
In this paper, we assume a field \textsf{ARG} to be names of things representing themselves, hence 
for example $\pi_{\textsf{ARG}}(\texttt{drug})$ is the set of names of drugs. 

For a value set $V$, we also consider inverse image 
$\pi^{-1}_{\textsf{N}}(V):=\{x\;|\; \pi_{\textsf{N}}(x)\in V\}$. For example, 
$$
\texttt{D}_1:=\pi^{-1}_{\textsf{SUBJ}}(\pi_{\textsf{ARG}}(\texttt{man}))
$$
consists of all tuples of the form $(\textsf{SUBJ}{=}x,\ldots)$, where $x$ is a man's name 
(i.e.~$x\in\pi_{\textsf{ARG}}(\texttt{man})$). 
Thus, $\texttt{sell}\cap\texttt{D}_1$ denotes men's selling events 
(i.e.~$\{(\textsf{SUBJ}{=}\textit{John}, \textsf{COMP}{=}\textit{Aspirin}),\ldots\}$ as in Figure~\ref{fig:dcswatch}). 
Similarly, the denotation of ``\textit{banned drugs}'' as in Figure~\ref{fig:dcstree}b is formally 
written as 
$$
\texttt{D}_2:=\texttt{drug}\cap\pi^{-1}_{\textsf{ARG}}(\pi_{\textsf{COMP}}(\texttt{ban})), 
$$
Hence the following denotation 
$$
\texttt{D}_3:=\texttt{sell}\cap\texttt{D}_1\cap\pi^{-1}_{\textsf{COMP}}(\pi_{\textsf{ARG}}(\texttt{D}_2))
$$
consists of selling events such that the \textsf{SUBJ} is a man and the \textsf{COMP} is a banned drug. 

The calculation above can proceed in a recursive manner controlled by DCS trees. 
The DCS tree for the sentence ``\textit{a man sells banned drugs}'' is shown in Figure~\ref{fig:dcswatch}. 
Formally, a DCS tree is defined as a rooted tree in which nodes are denotations of content words and 
edges are labeled by fields at each ends. Assume a node \texttt{x} has children $\texttt{y}_1,\ldots,\texttt{y}_n$, 
and the edges $(\texttt{x},\texttt{y}_1),\ldots,(\texttt{x},\texttt{y}_n)$ are labeled by 
$(\textsf{P}_1,\textsf{L}_1),\ldots,(\textsf{P}_n,\textsf{L}_n)$, respectively. Then, the denotation 
$\den{\texttt{x}}$ of the subtree rooted at \texttt{x} is recursively calculated as 
\begin{equation}
\label{eq:dendcs}
\den{\texttt{x}}:=\texttt{x}\cap\bigcap_{i=1}^{n}\pi^{-1}_{\textsf{P}_i}(\pi_{\textsf{L}_i}(\den{\texttt{y}_i})). 
\end{equation}
As a result, the denotation of the DCS tree in Figure~\ref{fig:dcswatch} is the denotation $\texttt{D}_3$ of 
``\textit{a man sells banned drugs}'' as calculated above. 
DCS can be further extended to handle phenomena such as quantifiers or superlatives \cite{liang13,tian14}. 
In this paper, we focus on the basic version, but note that it is already expressive enough to 
at least partially capture the meanings of a large portion of phrases and sentences. 

\begin{figure}[t]
\centering
\includegraphics[scale=0.26,bb=0 0 760 242,clip]{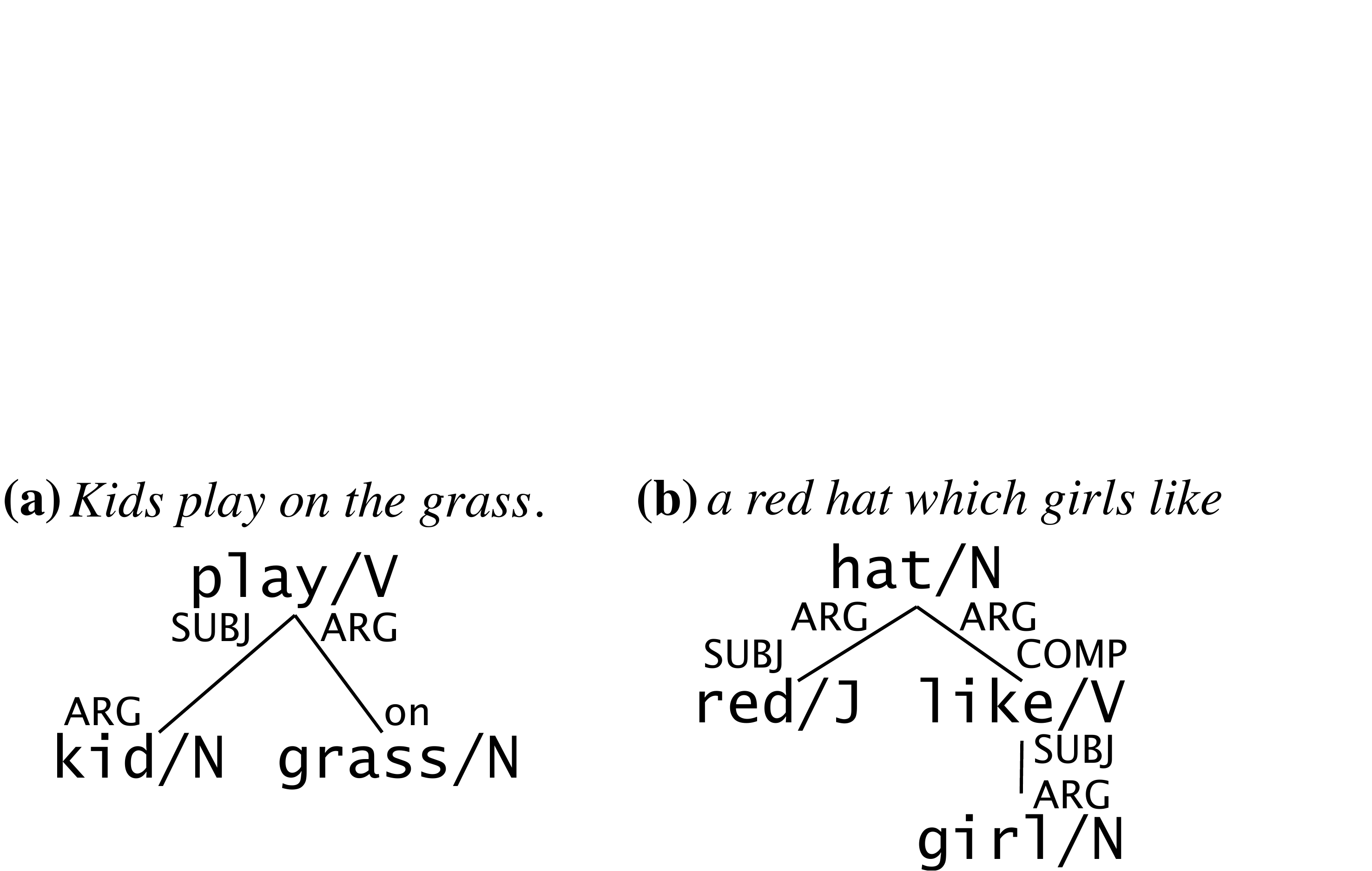}
\caption{DCS trees in this work}
\label{fig:dcsexamp}
\end{figure}

DCS trees can be learned from question-answer pairs and a given database of denotations \cite{liang13}, or they can 
be extracted from dependency trees if no database is specified, by taking advantage of the observation 
that DCS trees are similar to dependency trees \cite{tian14}. We use the latter approach, obtaining DCS trees by 
rule-based conversion from universal dependency (UD) trees \cite{mcdonald-EtAl:2013:Short}. 
Therefore, nodes in a DCS tree are 
content words in a UD tree, which are in the form of lemma-POS pairs (Figure~\ref{fig:dcsexamp}). The 
inventory of fields is designed to be \textsf{ARG}, \textsf{SUBJ}, \textsf{COMP}, and all prepositions. 
Prepositions are unlike content words which denote sets of things, but act as relations 
which we treat similarly as \textsf{SUBJ} and \textsf{COMP}. For example, a prepositional phrase attached 
to a verb (e.g.~\textit{play on the grass}) is treated as in Figure~\ref{fig:dcsexamp}a. 
The presence of two field labels on each edge of a DCS tree makes it convenient for modeling semantics in 
several cases, such as a relative clause (Figure~\ref{fig:dcsexamp}b). 

\section{Vector-based DCS}
\label{sec:vector}

For any content word \textit{w}, we use a query vector $\mathbf{v}_{\textit{w}}$ to model its denotation, 
and an answer vector $\mathbf{u}_{\textit{w}}$ to model a prototypical element in that denotation. Query 
vector $\mathbf{v}$ and answer vector $\mathbf{u}$ are learned such that $\exp(\mathbf{v}\cdot\mathbf{u})$ 
is proportional to the probability of $\mathbf{u}$ answering the query $\mathbf{v}$. The learning source 
is a collection of DCS trees, based on the idea that the DCS tree of a declarative sentence usually 
has non-empty denotation. 
For example, ``\textit{kids play}'' means there exists some kid who plays. 
Consequently, some element in the \texttt{play} denotation belongs to
$\pi^{-1}_{\textsf{SUBJ}}(\pi_{\textsf{ARG}}(\texttt{kid}))$, and some element in the \texttt{kid} 
denotation belongs to $\pi^{-1}_{\textsf{ARG}}(\pi_{\textsf{SUBJ}}(\texttt{play}))$. 
This is a signal to increase the dot product of  $\mathbf{u}_{\textit{play}}$ and the query vector 
of $\pi^{-1}_{\textsf{SUBJ}}(\pi_{\textsf{ARG}}(\texttt{kid}))$, as well as 
the dot product of  $\mathbf{u}_{\textit{kid}}$ and the query vector 
of $\pi^{-1}_{\textsf{ARG}}(\pi_{\textsf{SUBJ}}(\texttt{play}))$. When 
optimized on a large corpus, the ``typical'' elements of \texttt{play} and \texttt{kid} should be learned 
by $\mathbf{u}_{\textit{play}}$ and $\mathbf{u}_{\textit{kid}}$, respectively. 
In general, one has 
\begin{theorem}
Assume the denotation of a DCS tree is not empty. Given any path from node {\tt x} to {\tt y}, assume 
edges along the path are labeled by $({\sf P},{\sf L}),\ldots,({\sf K},{\sf N})$. Then, an 
element in the denotation {\tt y} belongs to 
$\pi^{-1}_{{\sf N}}(\pi_{{\sf K}}(\ldots(\pi^{-1}_{{\sf L}}(\pi_{{\sf P}}({\tt x})\ldots)$.
\end{theorem}
Therefore, for any two nodes in a DCS tree, the path from one to another forms a training example, 
which signals increasing the dot product of the corresponding query and answer vectors. 

It is noteworthy that the above formalization happens to be closely related to the skip-gram model \cite{word2vecNIPS}. The skip-gram learns a 
target vector $\mathbf{v}_{\textit{w}}$ and a context vector $\mathbf{u}_{\textit{w}}$ for each word \textit{w}. 
It assumes the 
probability of a word $\textit{y}$ co-occurring with a word $\textit{x}$ in a context window is proportional 
to $\exp(\mathbf{v}_{\textit{x}}\cdot\mathbf{u}_{\textit{y}})$. Hence, if $\textit{x}$ and $\textit{y}$ co-occur 
within a context window, then one gets a signal to increase $\mathbf{v}_{\textit{x}}\cdot\mathbf{u}_{\textit{y}}$. 
If the context window is taken as the same DCS tree, then the learning of skip-gram and vector-based DCS 
will be almost the same, except that 
%
%
%
%
the target vector $\mathbf{v}_{\textit{x}}$ becomes the query vector $\mathbf{v}$, which is 
no longer assigned to the word $\textit{x}$ but the path from $\textit{x}$ to $\textit{y}$ in the DCS tree 
(e.g.~the query vector for $\pi^{-1}_{\textsf{SUBJ}}(\pi_{\textsf{ARG}}(\texttt{kid}))$ instead of 
$\mathbf{v}_{\textit{kid}}$). Therefore, our model can also be regarded as extending skip-gram to 
take account of the changes of meanings caused by different syntactic-semantic roles. 

\paragraph{Additive Composition} Word vectors trained by skip-gram are known to be semantically additive, 
such as exhibited in word analogy tasks. An effect of adding up two skip-gram vectors is further analyzed in 
\newcite{addcomp}. 
Namely, the target vector $\mathbf{v}_{\textit{w}}$ can be regarded as encoding the distribution of context words 
surrounding \textit{w}. If another word \textit{x} is given, $\mathbf{v}_{\textit{w}}$ can 
be decomposed into two parts, one encodes context words shared with \textit{x}, and another encodes 
context words not shared. 
When $\mathbf{v}_{\textit{w}}$ and 
$\mathbf{v}_{\textit{x}}$ are added up, the non-shared part of each of them tend to cancel out, 
because non-shared parts have nearly independent distributions. 
As a result,  the shared part gets reinforced. 
An error bound is derived to estimate how close $\frac{1}{2}(\mathbf{v}_{\textit{w}}+\mathbf{v}_{\textit{x}})$ 
gets to the distribution of the shared part. 
We can see the 
%
same mechanism exists in vector-based DCS. In a DCS tree, 
two paths share a context word if they lead to a same node \textit{y}; semantically, this means some element 
in the denotation \texttt{y} belongs to both denotations of the two paths (e.g.~given the sentence 
``\textit{kids play balls}'', $\pi^{-1}_{\textsf{SUBJ}}(\pi_{\textsf{ARG}}(\texttt{kid}))$ and 
$\pi^{-1}_{\textsf{COMP}}(\pi_{\textsf{ARG}}(\texttt{ball}))$ both contain a playing event whose 
\textsf{SUBJ} is a kid and \textsf{COMP} is a ball). Therefore, addition of query vectors of two paths 
approximates their intersection because the shared context \textit{y} gets reinforced. 

\paragraph{Projection} Generally, for any two denotations $\texttt{X}_1,\texttt{X}_2$ and any projection 
$\pi_{\textsf{N}}$, we have 
\begin{equation}
\label{eq:piinter}
\pi_{\textsf{N}}(\texttt{X}_1\cap\texttt{X}_2)\subseteq\pi_{\textsf{N}}(\texttt{X}_1)\cap\pi_{\textsf{N}}(\texttt{X}_2). 
\end{equation}
And the ``$\subseteq$'' can often become ``$=$'', for example when $\pi_{\textsf{N}}$ is a one-to-one map or 
$\texttt{X}_1=\pi^{-1}_{\textsf{N}}(V)$ for some value set $V$. Therefore, if 
intersection is realized by addition, it will be natural to realize projection by linear mapping 
because  
\begin{equation}
\label{eq:linear}
(\mathbf{v}_{1}+\mathbf{v}_{2})M_{\textsf{N}}=\mathbf{v}_{1}M_{\textsf{N}}+\mathbf{v}_{2}M_{\textsf{N}}
\end{equation}
holds for any vectors $\mathbf{v}_{1},\mathbf{v}_{2}$ and any matrix 
$M_{\textsf{N}}$, which is parallel to \eqref{eq:piinter}. 
If $\pi_{\textsf{N}}$ is realized by a matrix $M_{\textsf{N}}$, then $\pi^{-1}_{\textsf{N}}$ should correspond to 
the inverse matrix $M^{-1}_{\textsf{N}}$, because 
$\pi_{\textsf{N}}(\pi^{-1}_{\textsf{N}}(V))=V$ for any value set $V$. So we have realized all composition operations 
in DCS. 

\paragraph{Query vector of a DCS tree} 
Now, we can define the query vector of a DCS tree as parallel to \eqref{eq:dendcs}:
\begin{equation}
\label{eq:querydcs}
\mathbf{v}_{\den{\texttt{x}}}:=\mathbf{v}_{\textit{x}}+\frac{1}{n}\sum_{i=1}^{n}
\mathbf{v}_{\den{\texttt{y}_i}}M_{\textsf{L}_i}M^{-1}_{\textsf{P}_i}. 
\end{equation}

\section{Training}
\label{sec:model}

As described in Section~\ref{sec:vector}, vector-based DCS 
assigns 
a query vector $\mathbf{v}_{\textit{w}}$ and an answer vector $\mathbf{u}_{\textit{w}}$ to each content 
word \textit{w}. And for each field \textsf{N}, it assigns two matrices $M_{\textsf{N}}$ and $M^{-1}_{\textsf{N}}$. 
For any path from node \textit{x} to \textit{y} sampled from a DCS tree, assume the 
edges along are labeled by $({\sf P},{\sf L}),\ldots,({\sf K},{\sf N})$. Then, the dot product 
$\mathbf{v}_{\textit{x}}M_{\textsf{P}}M^{-1}_{\textsf{L}}\ldots M_{\textsf{K}}M^{-1}_{\textsf{N}}\cdot\mathbf{u}_{\textit{y}}$ gets a signal to increase. 

Formally, we adopt the noise-contrastive estimation \cite{gutmann12} as used in the skip-gram model, and mix 
the paths sampled from DCS trees with artificially generated noise. Then, 
$\sigma(\mathbf{v}_{\textit{x}}M_{\textsf{P}}M^{-1}_{\textsf{L}}\ldots M_{\textsf{K}}M^{-1}_{\textsf{N}}\cdot\mathbf{u}_{\textit{y}})$ models 
the probability of a training example coming from DCS trees, where $\sigma(\theta)=1/\{1+\exp(-\theta)\}$ is 
the sigmoid function. 
The vectors and matrices are trained by maximizing the log-likelihood of the mixed data. 
We use stochastic gradient 
descent \cite{bottou12} for training. Some important settings are discussed below. 

\paragraph{Noise} For any $\mathbf{v}_{\textit{x}}M_{1}M^{-1}_{2}\ldots M_{2l-1}M^{-1}_{2l}\cdot\mathbf{u}_{\textit{y}}$ obtained from a path of a DCS tree, we 
generate noise 
by randomly choosing an index $i\in [2,2l]$, and then replacing $M_j$ or $M^{-1}_j$ ($\forall j\geq i$) and 
$\mathbf{u}_{\textit{y}}$ by $M_{\textsf{N}(j)}$ or $M^{-1}_{\textsf{N}(j)}$ and $\mathbf{u}_{\textit{z}}$, 
respectively, where $\textsf{N}(j)$ and \textit{z} are independently drawn from the marginal (i.e.~unigram) 
distributions of fields and words. 

\paragraph{Update} For each data point, when $i$ is the chosen index above 
for generating noise, we view indices $j<i$ as the "target" part, and $j>=i$ as the "context", which is 
completely replaced by the noise, as an analogous to the skip-gram model. Then, at each step we 
only update one vector and one matrix from each of the target, context, and noise part; more specifically, 
we only update $\mathbf{v}_{\textit{x}}$, $M_{i-1}$ or $M^{-1}_{i-1}$, $M_{i}$ or $M^{-1}_{i}$, 
$M_{\textsf{N}(i)}$ or $M^{-1}_{\textsf{N}(i)}$, $\mathbf{u}_{\textit{y}}$ and $\mathbf{u}_{\textit{z}}$, at the 
step. This is much faster than always updating all matrices. 

\paragraph{Initialization} Matrices are initialized as $\frac{1}{2}(I+G)$, where $I$ is the identity matrix; and 
$G$ and all vectors are initialized with i.i.d. Gaussians of variance $1/d$, where $d$ is the vector 
dimension. We find that the diagonal component $I$ is necessary to bring information from 
$\mathbf{v}_{\textit{x}}$ to $\mathbf{u}_{\textit{y}}$, whereas the randomness of $G$ makes convergence faster. 
$M^{-1}_{\textsf{N}}$ is initialized as the transpose of $M_{\textsf{N}}$. 

\paragraph{Learning Rate} We find that the initial learning rate for vectors can be set to $0.1$. 
But for matrices, it should be less than $0.0005$ otherwise the model diverges. 
For stable training, we rescale gradients when their norms exceed a threshold. 

\paragraph{Regularizer} During training, $M_{\textsf{N}}$ and $M^{-1}_{\textsf{N}}$ are treated as independent 
matrices. However, we use the regularizer 
$\gamma\lVert M^{-1}_{\textsf{N}}M_{\textsf{N}}-\frac{1}{d}\tr(M^{-1}_{\textsf{N}}M_{\textsf{N}})I \rVert^2$ to 
drive $M^{-1}_{\textsf{N}}$ close to the inverse of $M_{\textsf{N}}$.\footnote{Problem with the naive regularizer 
$\lVert M^{-1}M - I \rVert^2$ is that, when the scale of $M$ goes larger, it will drive 
$M^{-1}$ smaller, which may lead to degeneration. So we scale $I$ according to the trace of $M^{-1}M$.}
 We also use 
$\kappa\lVert M^{\bot}_{\textsf{N}}M_{\textsf{N}}-\frac{1}{d}\tr(M^{\bot}_{\textsf{N}}M_{\textsf{N}})I \rVert^2$ to 
prevent $M_{\textsf{N}}$ from having too different scales at different directions (i.e., to drive $M_{\textsf{N}}$ 
close to orthogonal). 
We set $\gamma=0.001$ and $\kappa=0.0001$. 
Despite the rather weak regularizer, we find that $M^{-1}_{\textsf{N}}$ can be learned to be exactly the inverse of 
$M_{\textsf{N}}$, and $M_{\textsf{N}}$ can actually be an orthogonal matrix, showing some semantic regularity 
(Section~\ref{sec:quality}). 

\begin{table}[t]
\centering
\scriptsize
\setlength{\tabcolsep}{3pt}
\it
\begin{tabular}{ | c | c | c | c |}
\hline
 {\rm GloVe} & {\rm no matrix} & {\rm vecDCS} & {\rm vecUD}\\
\hline
books & essay/N & novel/N & essay/N\\
author & novel/N & essay/N & novel/N\\
\textbf{published} & memoir/N & anthology/N & article/N\\
novel & books/N & publication/N & anthology/N\\
memoir & autobiography/N & memoir/N & poem/N\\
\textbf{wrote} & \textbf{non-fiction/J} & poem/N & autobiography/N\\
biography & \textbf{reprint/V} & autobiography/N & publication/N\\
autobiography & \textbf{publish/V} & story/N & journal/N\\
essay & \textbf{republish/V} & pamphlet/N & memoir/N\\
\textbf{illustrated} & chapbook/N & tale/N & pamphlet/N\\
\hline
\end{tabular}
\rm
\caption{Top 10 similar words to ``\textit{book/N}''}
\label{tab:simwords}
\end{table}

\section{Experiments}
\label{sec:experiments}

For training vector-based DCS, we use Wikipedia Extractor\footnote{\url{http://medialab.di.unipi.it/wiki/Wikipedia_Extractor}} to extract texts from 
the 2015-12-01 dump of English 
Wikipedia\footnote{\url{https://dumps.wikimedia.org/enwiki/}}. Then, we use Stanford 
Parser\footnote{\url{http://nlp.stanford.edu/software/lex-parser.shtml}} \cite{stanfordFactorizedParser} to parse 
all sentences and convert the UD trees into DCS trees by handwritten rules. We assign a weight to each path 
of the DCS trees as follows. 

\begin{table}[t]
\centering
\scriptsize
\renewcommand{\arraystretch}{1.15}
\setlength{\tabcolsep}{3pt}
\it
\begin{tabular}{ | c | c | c | }
\hline
$\pi^{-1}_{{\sf SUBJ}}(\pi_{{\sf ARG}}({\tt house}))$ & $\pi^{-1}_{{\sf COMP}}(\pi_{{\sf ARG}}({\tt house}))$ & $\pi^{-1}_{{\sf ARG}}(\pi_{{\sf in}}({\tt house}))$\\
\hline
victorian/J & build/V & sit/V\\
stand/V & rent/V & house/N\\
vacant/J & leave/V & stand/V\\
18th-century/J & burn down/V & live/V\\
historic/J & remodel/V & hang/V\\
old/J & demolish/V & seat/N\\
georgian/J & restore/V & stay/V\\
local/J & renovate/V & serve/V\\
19th-century/J & rebuild/V & reside/V\\
tenement/J & construct/V & hold/V\\
\hline
$\pi^{-1}_{{\sf ARG}}(\pi_{{\sf SUBJ}}({\tt learn}))$ & $\pi^{-1}_{{\sf ARG}}(\pi_{{\sf COMP}}({\tt learn}))$ & $\pi^{-1}_{{\sf about}}(\pi_{{\sf ARG}}({\tt learn}))$\\
\hline
teacher/N & skill/N & otherness/N\\
skill/N & lesson/N & intimacy/N\\
he/P & technique/N & femininity/N\\
she/P & experience/N & self-awareness/N\\
therapist/N & ability/N & life/N\\
student/N & something/N & self-expression/N\\
they/P & knowledge/N & sadomasochism/N\\
mother/N & language/N & emptiness/N\\
lesson/N & opportunity/N & criminality/N\\
father/N & instruction/N & masculinity/N\\
\hline
\end{tabular}
\rm
\caption{Top 10 answers of high dot products}
\label{tab:mattrans}
\end{table}

For any path $P$ passing through $k$ intermediate nodes of degrees $n_1,\ldots,n_k$, respectively, we 
set 
\begin{equation}
\label{eq:weightpath}
{\rm Weight}(P):=\prod_{i=1}^{k}\frac{1}{n_i - 1}.
\end{equation}
Note that $n_i\geq 2$ because there is a path $P$ passing through the node; and 
${\rm Weight}(P)=1$ if $P$ consists of a single edge. 
The equation \eqref{eq:weightpath} is intended to degrade long paths which pass through several high-valency 
nodes.
We use a random walk algorithm to sample paths such that the expected times a path is sampled equals its weight. 
As a result, the sampled path lengths range from $1$ to $19$, average $2.1$, with an exponential tail. 
We convert all words which are sampled less than 1000 times to \texttt{*UNKNOWN*/POS}, and all prepositions 
occurring less than 10000 times to an \textsf{*UNKNOWN*} field. As a result, we obtain a vocabulary of 
109k words and 211 field names. 

\begin{table*}[t]
\centering
\footnotesize
\setlength{\tabcolsep}{10pt}
\begin{tabular}{ l | c c c c c c }
\hline
 & AN & NN & VO & SVO & GS11 & GS12\\
 \hline
vecDCS & \textbf{0.51} & \textbf{0.49} & \textbf{0.41} & \textbf{0.62} & 0.29 & 0.33\\
 \;\;-no matrix & \textbf{0.52} & 0.46 & \textbf{0.42} & \textbf{0.62} & 0.29 & 0.33\\
 \;\;-no inverse & 0.47 & 0.43 & 0.38 & 0.58 & 0.28 & 0.33\\
vecUD & 0.44 & 0.46 & \textbf{0.41} & 0.58 & 0.25 & 0.25\\
GloVe & 0.41 & 0.47 & \textbf{0.41} & 0.60 & 0.23 & 0.17\\
\hline
\newcite{grefenstette-sadrzadeh:2011:EMNLP} & - & - & - & - & 0.21 & - \\
\newcite{blacoe-lapata:2012:EMNLP-CoNLL}:RAE & 0.31 & 0.30 & 0.28 & - & - & -\\
\newcite{Grefenstette:PhDthesis} & - & - & - & - & - & 0.27\\
\newcite{paperno-pham-baroni14} & - & - & - & - & - & \textbf{0.36}\\
\newcite{hashimoto-EtAl:2014:EMNLP2014}:$\text{Wadd}_{\text{nl}}$ & 0.48 & 0.40 & 0.39 &  & 0.34 & -\\
\newcite{kartsadrqpl2014} & - & - & - & 0.43 & \textbf{0.41} & -\\
\hline
\end{tabular}
\caption{Spearman's $\rho$ on phrase similarity}
\label{tab:phrasesim}
\end{table*}

Using the sampled paths, vectors and matrices are trained as in Section~\ref{sec:model} (vecDCS). 
The vector dimension is set to $d=250$. 
We compare with three baselines: (i) all matrices are fixed to identity (``no matrix''), in order 
to investigate the effects of meaning changes caused by syntactic-semantic roles and prepositions; (ii) the 
regularizer enforcing $M^{-1}_{\textsf{N}}$ to be actually the inverse matrix of $M_{\textsf{N}}$ is set to 
$\gamma=0$ (``no inverse''), in order to investigate the effects of a semantically motivated constraint; and 
(iii) applying the same training scheme to UD trees directly, by modeling UD relations as matrices 
(``vecUD''). In this case, one edge is assigned one UD relation \textsf{rel}, 
so we implement the transformation from child to 
parent by $M_{\textsf{rel}}$, and from parent to child by $M^{-1}_{\textsf{rel}}$. 
The same hyper-parameters are used to train vecUD. By comparing vecDCS with vecUD we investigate 
if applying the semantics framework of DCS makes any difference. 
Additionally, we compare with the 
GloVe (6B, 300d) vector\footnote{\url{http://nlp.stanford.edu/projects/glove/}} 
\cite{pennington-socher-manning:2014:EMNLP2014}. Norms of all word vectors are normalized to 1 
and Frobenius norms of all matrices are normalized to $\sqrt{d}$. 

\subsection{Qualitative Analysis}
\label{sec:quality}

We observe several special properties of the vectors and matrices trained by our model. 

\paragraph{Words are clustered by POS} In terms of cosine similarity, word vectors trained by vecDCS and vecUD are clustered by 
POS tags, probably due to their interactions with 
matrices during training. 
This is in contrast 
to the vectors trained by GloVe or ``no matrix'' (Table~\ref{tab:simwords}). 

\paragraph{Matrices show semantic regularity} Matrices learned for \textsf{ARG}, \textsf{SUBJ} and \textsf{COMP} 
are exactly orthogonal, and some most frequent prepositions\footnote{\textsf{of}, \textsf{in}, \textsf{to}, \textsf{for}, \textsf{with}, 
\textsf{on}, \textsf{as}, \textsf{at}, \textsf{from}} 
are remarkably close. 
For these matrices, the corresponding $M^{-1}$ also exactly converge to their inverse. 
It suggests regularities in the semantic space, especially because orthogonal matrices preserve cosine 
similarity -- if $M_{\textsf{N}}$ is orthogonal, two words \texttt{x}, \texttt{y} and their projections 
$\pi_{\textsf{N}}(\texttt{x})$, $\pi_{\textsf{N}}(\texttt{y})$ will have the same similarity measure, which 
is semantically reasonable. In contrast, matrices trained by vecUD are only orthogonal for three UD relations, 
namely \textsf{conj}, \textsf{dep} and \textsf{appos}. 


\paragraph{Words transformed by matrices} To illustrate the matrices trained by vecDCS, we start from the 
query vectors of two words, \texttt{house} and \texttt{learn}, applying different matrices to them, and show the 
10 answer vectors of the highest dot products (Tabel~\ref{tab:mattrans}). These are the lists of likely 
words which: take \textit{house} as a subject, take \textit{house} as a complement, fills into 
``\textit{$\rule{2ex}{1pt}$ in house}'', serve as a subject of \textit{learn}, serve as a complement of \textit{learn}, 
and fills into ``\textit{learn about $\rule{2ex}{1pt}$}'', respectively. As the table shows, matrices in 
vecDCS are appropriately 
learned to map word vectors to their syntactic-semantic roles. 

\begin{table*}[t]
\centering
\scriptsize
\it
\begin{tabular}{ | l | l | }
\hline
 {\rm Message-Topic($e_1$, $e_2$)} & It is a monthly $[\text{report}]_1$ providing $[\text{opinion}]_2$ and advice on current United States government contract issues. \\
\hline
{\rm Message-Topic($e_1$, $e_2$)} & The $[\text{report}]_1$ gives an account of the silvicultural $[\text{work}]_2$ done in Africa, Asia, Australia, South American and the Caribbean.\\
{\rm Message-Topic($e_1$, $e_2$)} & NUS today responded to the Government's $[\text{announcement}]_1$ of the long-awaited 
$[\text{review}]_2$ of university funding.\\
{\rm Component-Whole($e_2$, $e_1$)} & The $[\text{review}]_1$ published political $[\text{commentary}]_2$ and opinion, but even more than that.\\
{\rm Message-Topic($e_1$, $e_2$)} & It is a 2004 $[\text{book}]_1$ criticizing the political and linguistic $[\text{writings}]_2$ of Noam Chomsky.\\
\hline
\end{tabular}
\rm
\caption{Similar training instances clustered by cosine similarities between features}
\label{tab:relexamp}
\end{table*}

\subsection{Phrase Similarity}
\label{sec:phrasesim}

To test if vecDCS has the composition ability to calculate similar things as similar vectors, we conduct evaluation 
on a wide range of phrase similarity tasks. In these tasks, a system calculates similarity scores for pairs of 
phrases, and the performance is evaluated as its correlation 
with human annotators, measured by Spearman's $\rho$. 

\paragraph{Datasets} \newcite{mitchell10} create datasets\footnote{\url{http://homepages.inf.ed.ac.uk/s0453356/}}  for pairs of three types of two-word phrases: 
adjective-nouns (AN) (e.g.~``\textit{black hair}'' and ``\textit{dark eye}''), compound nouns (NN) (e.g.~``\textit{tax charge}'' and ``\textit{interest rate}'') and verb-objects (VO) (e.g.~``\textit{fight war}'' and ``\textit{win battle}''). 
Each dataset 
consists of 108 pairs and each pair is annotated by 18 humans (i.e., 1,944 scores in total). 
Similarity scores are integers ranging from 1 to 7. 
Another dataset\footnote{\url{http://www.cs.ox.ac.uk/activities/compdistmeaning/}} is created by extending VO to 
Subject-Verb-Object (SVO), and then assessing similarities by crowd sourcing \cite{kartsadrqpl2014}. 
The dataset GS11 created by \newcite{grefenstette-sadrzadeh:2011:EMNLP} (100 pairs, 25 annotators) is also of 
the form SVO, but 
in each pair only the verbs are different (e.g.~``\textit{man provide/supply money}''). The dataset GS12 
 described in \newcite{Grefenstette:PhDthesis} (194 pairs, 50 annotators) 
is of the form Adjective-Noun-Verb-Adjective-Noun (e.g.~``\textit{local family run/move small hotel}''), 
where only verbs are different in each pair. 

\paragraph{Our method} We calculate the cosine similarity of query vectors corresponding to phrases. For 
example, the query vector for ``\textit{fight war}'' is calculated as 
$\mathbf{v}_{\textit{war}}M_{\textsf{ARG}}M^{-1}_{\textsf{COMP}}+\mathbf{v}_{\textit{fight}}$. 
For vecUD we use $M_{\textsf{nsubj}}$ and $M_{\textsf{dobj}}$ instead of $M_{\textsf{SUBJ}}$ 
and $M_{\textsf{COMP}}$, respectively. For GloVe we use additive compositions. 

\begin{figure}[t]
\centering
\includegraphics[scale=0.26,bb=0 0 680 360,clip]{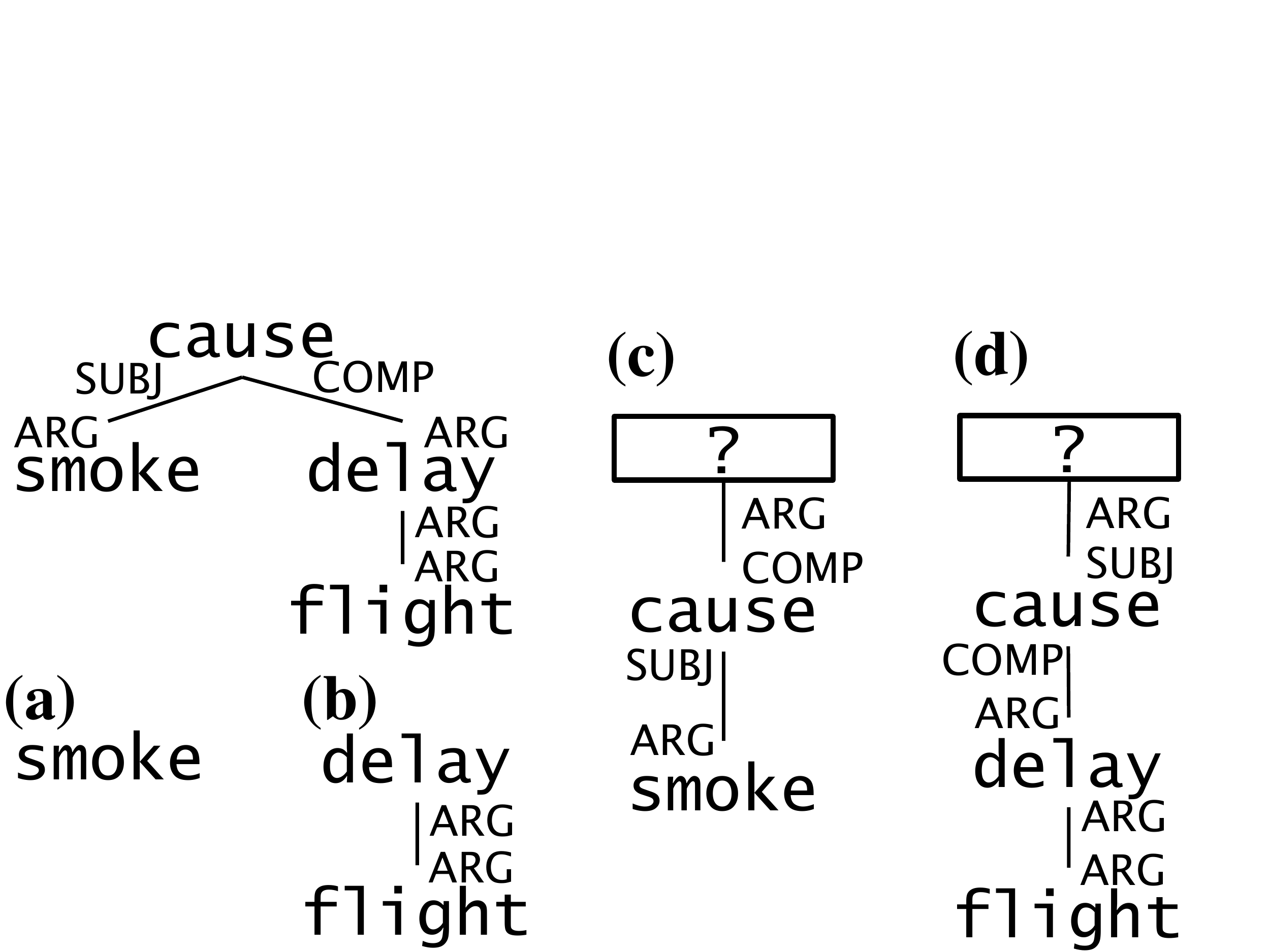}
\caption{For ``\textit{$[\text{smoke}]_1$ cause flight $[\text{delay}]_2$}'', we construct (a)(b) 
from subtrees, and (c)(d) from re-rooted trees, to form 4 query vectors as feature.} 
\label{fig:fourvec}
\end{figure}

\paragraph{Results} As shown in Table~\ref{tab:phrasesim}, vecDCS is competitive 
on AN, NN, VO, SVO and GS12,  consistently outperforming ``no inverse'', vecUD and GloVe, showing 
strong compositionality. 
The weakness of ``no inverse'' suggests that relaxing the constraint of inverse matrices may hurt compositionaly, 
though our preliminary examination on word similarities did not find any difference. 
The GS11 dataset appears to favor models that can learn from interactions between the subject and 
object arguments, such as the non-linear model $\text{Wadd}_{\text{nl}}$ in \newcite{hashimoto-EtAl:2014:EMNLP2014} and the entanglement model in \newcite{kartsadrqpl2014}. However, 
these models do not show particular advantages on other datasets. 
The recursive autoencoder (RAE) proposed in \newcite{socher11} shares an aspect with vecDCS as 
to construct meanings from parse trees. It is tested by \newcite{blacoe-lapata:2012:EMNLP-CoNLL} 
for compositionality, where vecDCS appears to be better. 
Nevertheless, we note that ``no matrix'' performs as good as vecDCS, suggesting that 
meaning changes caused by syntactic-semantic roles might not be major factors in these datasets, 
because the syntactic-semantic relations are all fixed in each dataset. 

\subsection{Relation Classification}
\label{sec:relclass}

In a relation classification task, the relation between two words in a sentence needs to be classified; we 
expect vecDCS to perform better than ``no matrix'' on this task because vecDCS can distinguish the different 
syntactic-semantic roles of the two slots the two words fit in. We confirm this conjecture in this section. 

\begin{table}[t]
\centering
\footnotesize
\setlength{\tabcolsep}{12pt}
\begin{tabular}{ l | c }
\hline
vecDCS & 81.2 \\
 \;\;-no matrix & 69.2\\
 \;\;-no inverse & 79.7\\
vecUD & 69.2\\
GloVe & 74.1\\
\hline
\newcite{socher12} & 79.1\\
 \;\;+3 features & 82.4\\
\newcite{dossantos-xiang-zhou:2015:ACL-IJCNLP} & 84.1\\
\newcite{xu-EtAl:2015:EMNLP1} & \textbf{85.6}\\
\hline
\end{tabular}
\caption{F1 on relation classification}
\label{tab:relclass}
\end{table}

\paragraph{Dataset} We use the dataset of SemEval-2010 Task 8 \cite{hendrickx-EtAl:2009:SEW}, in which 9 directed relations 
(e.g.~\textit{Cause-Effect}) and 1 undirected relation \textit{Other} are annotated, 8,000 instances for 
training and 2,717 for test. Performance is measured by the 9-class direction-aware Macro-F1 score 
excluding \textit{Other} class. 

\paragraph{Our method} For any sentence with two words marked as $e_1$ and $e_2$, we construct 
the DCS tree of the sentence, and take the subtree $T$ rooted at the common ancestor of $e_1$ and $e_2$. 
We construct four vectors from $T$, namely: the query vector for the subtree rooted at $e_1$ (resp. $e_2$), 
and the query vector of the DCS tree obtained from $T$ by rerooting it at $e_1$ (resp. $e_2$) 
(Figure~\ref{fig:fourvec}). The 
four vectors are normalized and concatenated to form the only feature used to train a classifier. 
For vecUD, we use the corresponding vectors calculated from UD trees. For GloVe, we use the word vector 
of $e_1$ (resp. $e_2$), and the sum of vectors of  all words within the span 
$[e_1,e_2)$ (resp. $(e_1, e_2]$) as the four vectors. Classifier is 
SVM\footnote{\url{https://www.csie.ntu.edu.tw/~cjlin/libsvm/}} with RBF kernel, $C=2$ and $\Gamma=0.25$. 
The hyper-parameters are selected by 5-fold cross validation.

\paragraph{Results} VecDCS outperforms baselines on relation classification (Table~\ref{tab:relclass}). 
It makes 16 errors in misclassifying the direction of a relation, as compared to 
144 such errors made by ``no matrix'', 23 by ``no inverse'', 30 by vecUD, and 161 by GloVe. This suggests 
that models with syntactic-semantic transformations (i.e.~vecDCS, ``no inverse'', and vecUD) are indeed 
good at distinguishing the different roles played by $e_1$ and $e_2$. VecDCS scores moderately lower 
than the state-of-the-art \cite{xu-EtAl:2015:EMNLP1}, however we note that these results are achieved 
by adding additional features and training task-specific neural networks 
\cite{dossantos-xiang-zhou:2015:ACL-IJCNLP,xu-EtAl:2015:EMNLP1}. Our method only uses features 
constructed from unlabeled corpora. From this point of view, it is comparable to the MV-RNN model 
(without features) in \newcite{socher12}, and vecDCS actually does better. 
Table~\ref{tab:relexamp} shows an example of clustered training instances as assessed by 
cosine similarities between 
their features. It suggests that the features used in our method can actually cluster similar relations. 

\begin{table}[t]
\centering
\scriptsize
\it
\begin{tabular}{ | c | c | c | }
\hline
``banned drugs'' & ``banned movies'' & ``banned books''\\
\hline
drug/N & bratz/N & publish/N\\
marijuana/N & porn/N & unfair/N\\
cannabis/N & indecent/N & obscene/N\\
trafficking/N & blockbuster/N & samizdat/N\\
thalidomide/N & movie/N & book/N\\
smoking/N & idiots/N & responsum/N\\
narcotic/N & blacklist/N & illegal/N\\
botox/N & grindhouse/N & reclaiming/N\\
doping/N & doraemon/N & redbook/N\\
\hline
\end{tabular}
\rm
\caption{Answers for composed query vectors}
\label{tab:queryexamp}
\end{table}

\subsection{Sentence Completion}
\label{sec:sentcomp}

If vecDCS can compose query vectors of DCS trees, one should be able to ``execute'' the vectors 
to get a set of answers, as the original DCS trees can do. This is done by taking dot products with 
answer vectors and then ranking the answers. Examples are shown in Table~\ref{tab:queryexamp}. 
Since query vectors and answer vectors are trained 
from unlabeled corpora, we can only obtain a coarse-grained candidate list. However, it is noteworthy that 
despite a common word ``\textit{banned}'' shared by the phrases, their answer lists 
are largely different, suggesting that composition actually can be done. Moreover, some words indeed 
answer the queries (e.g.~\textit{Thalidomide} for ``banned drugs'' and \textit{Samizdat} for 
``\textit{banned books}''). 

Quantitatively, we evaluate this utility of executing queries on the sentence completion task. 
In this task, a sentence is presented 
with a blank that need to be filled in. Five possible words are given as options for each blank, and a 
system needs to choose the correct one. The task can be viewed as a coarse-grained question answering 
or an evaluation for language models \cite{zweig-EtAl:2012:ACL2012}. 
We use the MSR sentence completion dataset\footnote{\url{http://research.microsoft.com/en-us/projects/scc/}} 
which consists of 1,040 test questions and a corpus for training language models. We train vecDCS on this 
corpus and use it for evaluation. 

\paragraph{Results} As shown in Table~\ref{tab:sentcomp}, vecDCS scores better than the N-gram model 
and demonstrates promising performance. However, to our surprise, ``no matrix'' shows an even better 
result which is the new state-of-the-art. Here we might be facing the same problem as in the phrase 
similarity task (Section~\ref{sec:phrasesim}); namely, all choices in a question fill into the same blank and 
the same syntactic-semantic role, so the transforming matrices in vecDCS might not be able to distinguish 
different choices; on the other hand, vecDCS would suffer more from parsing and POS-tagging errors. Nonetheless, 
we believe the result by ``no matrix'' reveals a new horizon of sentence completion, 
and suggests that composing semantic vectors according to DCS trees could be a promising direction. 

\begin{table}[t]
\centering
\footnotesize
\begin{tabular}{ l | c }
\hline
vecDCS & 50\\
 \;\;-no matrix & \textbf{60}\\
 \;\;-no inverse & 46\\
vecUD & 31\\
\hline
N-gram (Various) & 39-41\\
\newcite{zweig-EtAl:2012:ACL2012} & 52\\
\newcite{Mnih12afast} & 55\\
\newcite{gubbins-vlachos:2013:EMNLP} & 50\\
\newcite{mikolovRNN} & 55\\
\hline
\end{tabular}
\caption{Accuracy (\%) on sentence completion}
\label{tab:sentcomp}
\end{table}

\section{Discussion}
\label{sec:discussion}

We have demonstrated a way to link a vector composition model to a formal semantics, combining the strength 
of vector representations to calculate phrase similarities, and the strength of formal semantics to build up 
structured queries. In this section, we discuss several lines of previous research related to this work. 

\paragraph{Logic and Distributional Semantics}

Logic is necessary for implementing the functional aspects of meaning and organizing knowledge in a 
structured and unambiguous way. In contrast, distributional semantics provides an elegant methodology for 
assessing semantic similarity and is well suited for learning from data. There have been repeated calls for 
combining the strength of these two approaches \cite{coecke10,baroni14frege,liangpotts:2014}, and several systems 
\cite{lewis13,beltagy-erk-mooney:2014:P14-1,tian14} have contributed to this 
direction. In the remarkable work by \newcite{beltagy:cl2016}, word and phrase similarities are explicitly transformed 
to weighted logical rules that are used in a probabilistic inference framework. However, 
this approach requires considerable amount of engineering, including the generation of rule candidates 
(e.g.~by aligning sentence fragments), converting distributional similarities to weights, and 
efficiently handling the rules and inference. 
What if the distributional representations are equipped with a logical interface, such that the inference can be 
realized by simple vector calculations? We have shown it possible to realize semantic composition; 
we believe this may lead to significant simplification of the system design for combining logic and distributional 
semantics. 

\paragraph{Compositional Distributional Models}

There has been active exploration on how to combine word vectors such that 
adequate phrase/sentence similarities can be assessed \cite[\emph{inter alia}]{mitchell10}, 
and there is nothing new in using matrices to model changes of meanings. However, 
previous model designs mostly rely on linguistic intuitions 
\cite[\emph{inter alia}]{paperno-pham-baroni14}, whereas our model has an exact logic interpretation. 
Furthermore, by using additive composition we enjoy a learning guarantee \cite{addcomp}. 

\paragraph{Vector-based Logic Models}

This work also shares the spirit with \newcite{grefenstette:2013:*SEM} and \newcite{rocktaschel14low}, 
in exploring vector calculations that realize 
logic operations. However, the previous works did not specify how to integrate contextual 
distributional information, which is necessary for calculating semantic similarity. 

\paragraph{Formal Semantics}

Our model implements a fragment of logic capable of semantic composition, largely due to the simple 
framework of Dependency-based Compositional Semantics \cite{liang13}. 
It fits in 
a long tradition of logic-based semantics 
\cite{montague:1970,intromontague1981,disclogic1993}, with extensive studies 
on extracting semantics from syntactic representations such as HPSG \cite{hpsg2001,hpsg2005} 
and CCG \cite{ccg2002,bos04,steedman12book,artzi-lee-zettlemoyer:2015:EMNLP,mineshima-EtAl:2015:EMNLP}. 

\paragraph{Logic for Natural Language Inference}

The pursue of a logic more suitable for natural
language inference is also not new. For example, \newcite{maccartney08} has implemented a
model of natural logic \cite{lakoff70}. We would not reach the current formalization of logic of DCS without 
reading the work by \newcite{calvanese98}, which is an elegant formalization of database semantics in description logic. 

\paragraph{Semantic Parsing} DCS-related representations 
have been actively used in semantic parsing and we see potential in applying our model. For example, 
\newcite{berant-liang:2014:P14-1} convert $\lambda$-DCS queries to 
canonical utterances and assess paraphrases at the surface level; an alternative could be using 
vector-based DCS to bring distributional similarity directly into calculation of denotations. We also 
borrow ideas from previous work, 
for example our training scheme is similar to \newcite{guu-miller-liang:2015:EMNLP} in using 
paths and composition of matrices, and our method is similar to \newcite{poon-domingos:2009:EMNLP} 
in building structured knowledge from clustering syntactic parse of unlabeled data.

\paragraph{Further Applications}

Regarding the usability of distributional representations learned by our model, a strong point is that the 
representation takes into account syntactic/structural information of context. Unlike several previous models 
\cite{padolapata07,levy-goldberg:2014:P14-2,pham-EtAl:2015:ACL-IJCNLP}, our approach learns matrices at the same time that can extract the 
information according to different syntactic-semantic roles. 
A related application is selectional preference \cite{Baroni:2010:DMG:1945043.1945049,lenci:2011:CMCL,vandecruys:2014:EMNLP2014}, 
wherein our model might has 
potential for smoothly handling composition. 

\paragraph{Reproducibility} Find our code at 
\url{https://github.com/tianran/vecdcs}

\paragraph{Acknowledgments}

This work was supported by CREST, JST. 
We thank the anonymous reviewers for their valuable comments.

\bibliography{ref}
\bibliographystyle{acl2016}

\end{document}